\documentclass{ifacconf}

\usepackage{graphicx}      
\usepackage{natbib}        
\usepackage{amssymb}
\usepackage{amsmath}
\usepackage{diagbox}
\usepackage{caption}
\usepackage{subcaption}
\usepackage{ragged2e}
\usepackage{float}
\usepackage[export]{adjustbox}
\usepackage{multirow}
\usepackage{booktabs}
\usepackage{tabularx}

\usepackage{xcolor}
\newcommand{\revise}[1]{\textcolor{black}{#1}}
\begin{document}
\begin{frontmatter}

\title{Vision-driven UAV River Following: Benchmarking with Safe Reinforcement Learning} 

\thanks[footnoteinfo]{This work was supported by ONR N00014-20-1-2085 and N00014-24-1-2019.}

\author[First]{Zihan Wang and Nina Mahmoudian} 

\address[First]{School of Mechanical Engineering, Purdue University, West Lafayette, IN 47907 USA (e-mail: wang5044,ninam@ purdue.edu).}

\begin{abstract}                
In this study, we conduct a comprehensive benchmark of the Safe Reinforcement Learning (Safe RL) algorithms for the task of vision-driven river following of Unmanned Aerial Vehicle (UAV) in a Unity-based photo-realistic simulation environment. 
We \revise{empirically} validate the effectiveness of \revise{semantic-augmented image encoding method}, assessing its superiority based on Relative Entropy and the quality of \revise{water} pixel reconstruction. The determination of the encoding \revise{dimension}, guided by reconstruction loss, contributes to a more compact state representation, facilitating the training of \revise{Safe RL} policies.
Across all benchmarked Safe RL algorithms, we find that First Order Constrained Optimization in Policy Space achieves the optimal balance between reward acquisition and \revise{safety compliance}.
Notably, our results reveal that on-policy algorithms consistently outperform both off-policy and model-based counterparts in both training and testing environments.
Importantly, the benchmarking outcomes and the vision encoding methodology extend beyond UAVs, \revise{and are} applicable to Autonomous Surface Vehicles (ASVs) engaged in \revise{autonomous navigation in confined waters}.
\end{abstract}


\end{frontmatter}

\section{Introduction}
Performing surveillance and rescue missions by autonomous vehicles in complex and unknown riverine environments requires safe control policy for unmanned navigation overcoming detrimental failures such as collision with natural or man-made obstacles. For example during self-driving navigation missions an ASV can run aground to river bank or sand island, and an UAV can bump with trees or bridges. 
Deep Reinforcement Learning (DRL) approaches has been gaining promising results in solving autonomous navigation tasks of mobile robots, \cite{wang2023deep,lee2022mobile, zhao2023autonomous}. 
Water segmentation mask have been used for vision-driven autonomous river following of an UAV in a relatively simple curved river channel, \cite{taufik2015multi, taufik2016multi}.
However, as illustrated in \cite{ray2019benchmarking, ji2023safety}, meaningful trade-offs tend to exist between task performance and safety objectives, so that the well-trained DRL model in unconstrained manner may not have safe behaviors automatically.
The safe policy can be achieved by injecting safety into DRL algorithms to satisfy safety constraints throughout exploration during training phase and inference during testing phase.
Safe RL bridges the RL simulation experiments to real-world applications, \cite{gu2022review}.

While DRL has proven effective in numerous game-like scenarios, its application to visual navigation within realistic environments presents a formidable challenge, \cite{kulhanek2019vision}. 
\cite{li2020unsupervised} tackles the visual navigation in a low-resource setting by an unsupervised RL approach, \cite{xiao2022multigoal} focuses on extracting collision information from visual observation to reshape reward  in navigating a ground agent, \cite{zielinski20213d} employs image processing module like object detection network to navigate an Autonomous Underwater Vehicle (AUV) to a target point, \cite{knyaz2020object} addresses the navigation of an UAV by jointly reconstructing 3D voxel model and semantic scene segmentation from images.   
Nevertheless, the exploration of DRL for vision-driven navigation in riverine domains, considering the interplay between task objectives and safety, remains a relatively under-researched and challenging area.
Thus it is critical to investigate how state-of-the-art Safe RL algorithms balance these trade-offs in the river following task, where the safe agent in this paper is focused on UAVs, but can be extended to ASVs as well.  


\revise{A riverine environment within Unreal Engine is constructed by \cite{liang2021vision} and \cite{wei2022vision} to train DRL algorithms for vision-driven UAV's autonomous river following task. However, safety considerations are not integrated into either the environment or the DRL algorithms.}
Existing benchmarking environments that incorporate safety constraints either focus on ground vehicles (\cite{ray2019benchmarking, ji2023safety, ji2023omnisafe, zhang2022saferl, xu2023benchmarking}), or non-vision observations as policy input for aerial vehicles (\cite{yuan2022safe}), few Safe RL environments target vision-driven control for autonomous navigation tasks \revise{in the riverine domain.}
In this regard, we present the \textbf{Safe Riverine Environment (SRE)}, \revise{an extension of our previous work, \cite{wang2024vision}, on photorealistic riverine simulation environment (RSE) for the river following task.
Compared to RSE's two maps, SRE} has three difficulty levels (\{\textit{Easy}, \textit{Medium}, \textit{Hard}\}, in accordance to environment structure design in \cite{ji2023safety}) with increasing task complexity. As environment difficulty increases, the number of bridges, river turns and sand islands, and the variations of river width and depth are increased. 
\revise{Safety constraints are firstly introduced in SRE, which are categorized into tight and loose constraints due to their different severity degrees.}
The fine-grained failure details in both constraint categories are provided \revise{in SRE} for more detailed statistical failure analysis and algorithm comparison.
\revise{Moreover, besides the pure RGB image input for vision encoding in RSE, SRE also accepts pure mask or RGB+Mask input for vision-driven navigation.}

\begin{figure}[h]
    \centering
    \captionsetup[subfigure]{justification=Centering}
    \begin{subfigure}[b]{0.15\textwidth}\label{fig:rse-easy}
        \includegraphics[height=2.5cm,width=\textwidth]{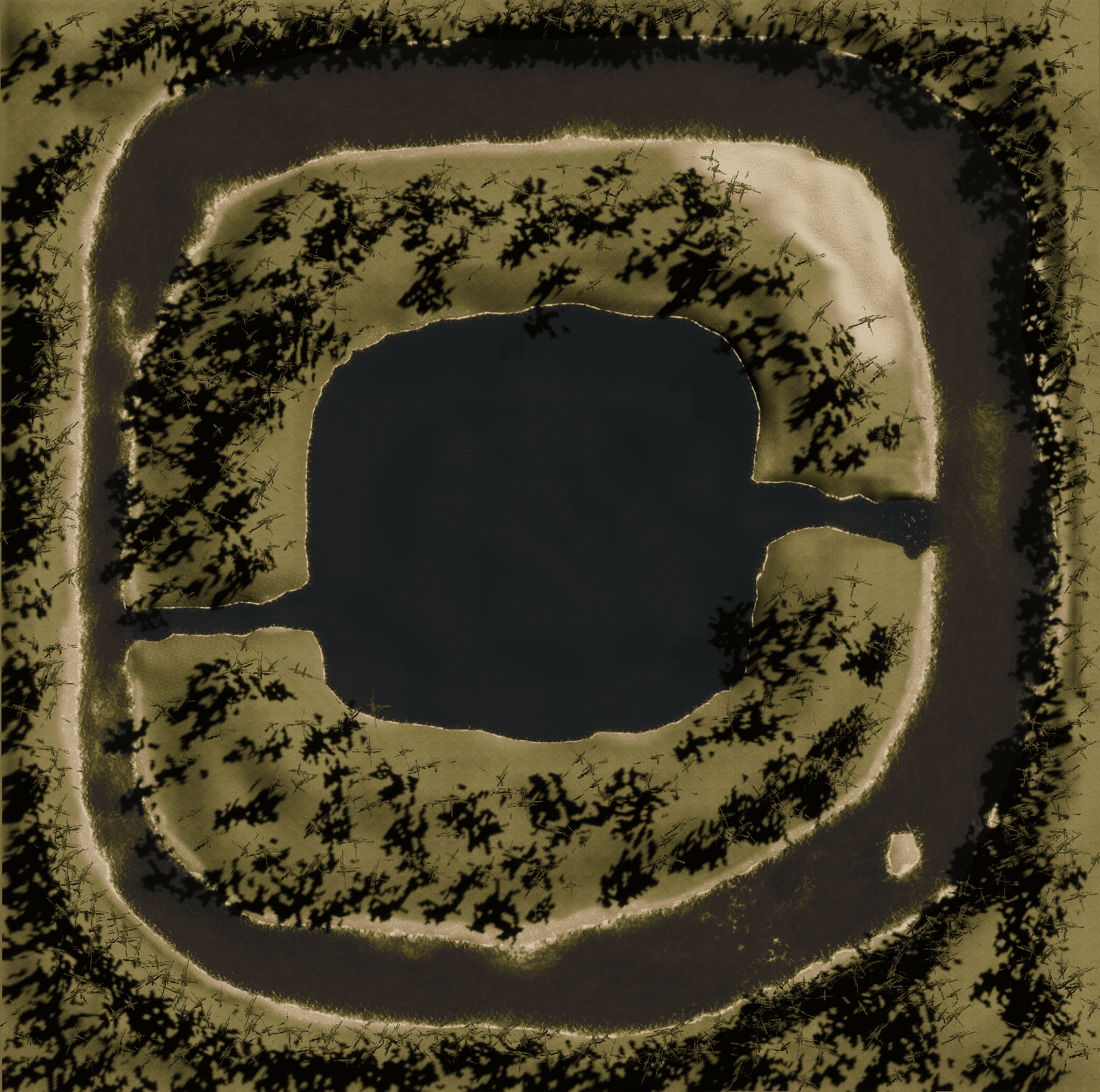}
        \caption{Easy}
    \end{subfigure}
    \begin{subfigure}[b]{0.15\textwidth}\label{fig:rse-medium}
        \includegraphics[height=2.5cm,width=\textwidth]{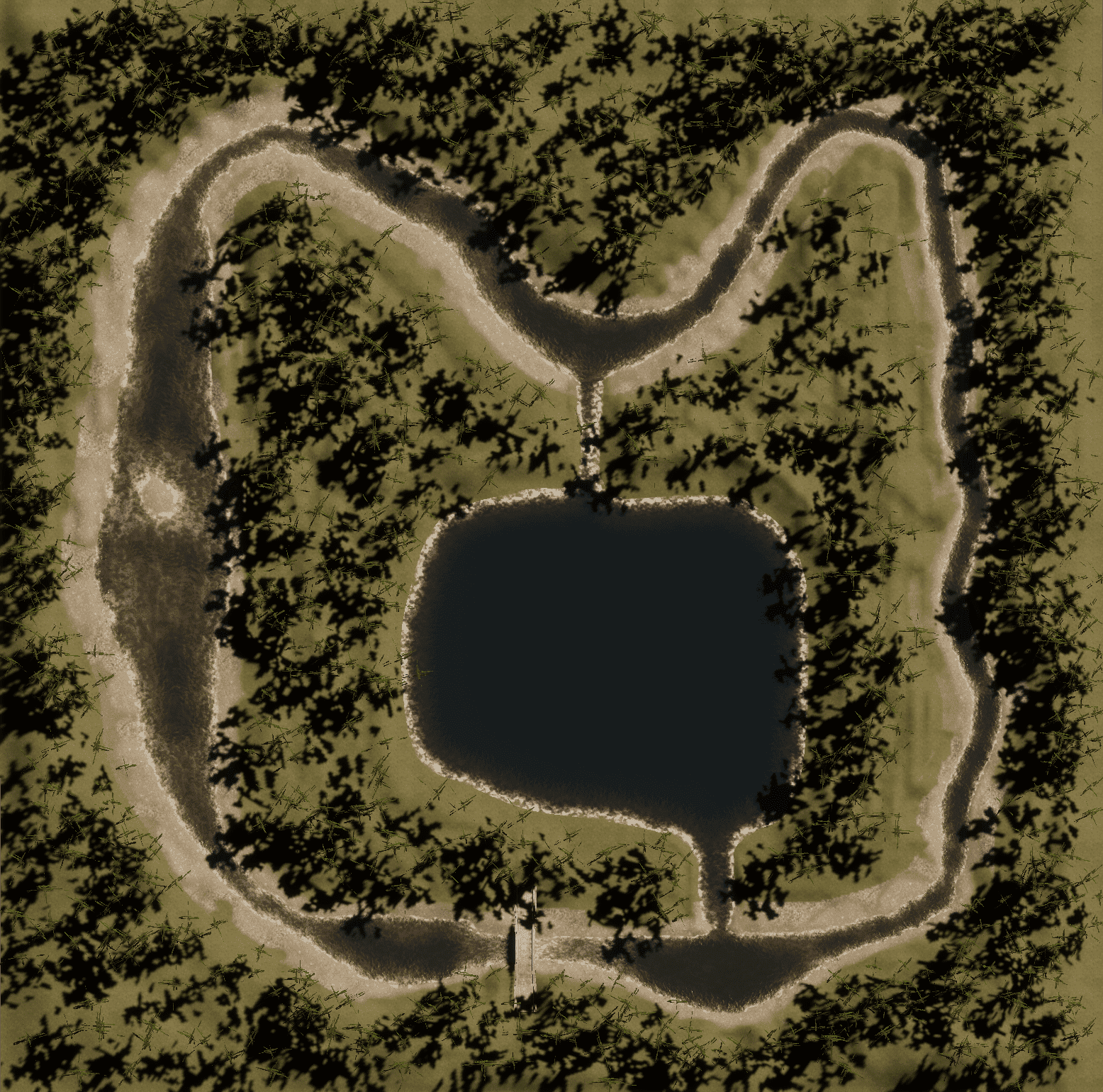}
        \caption{Medium}
    \end{subfigure}
    \begin{subfigure}[b]{0.15\textwidth}\label{fig:rse-hard}
        \includegraphics[height=2.5cm,width=\textwidth]{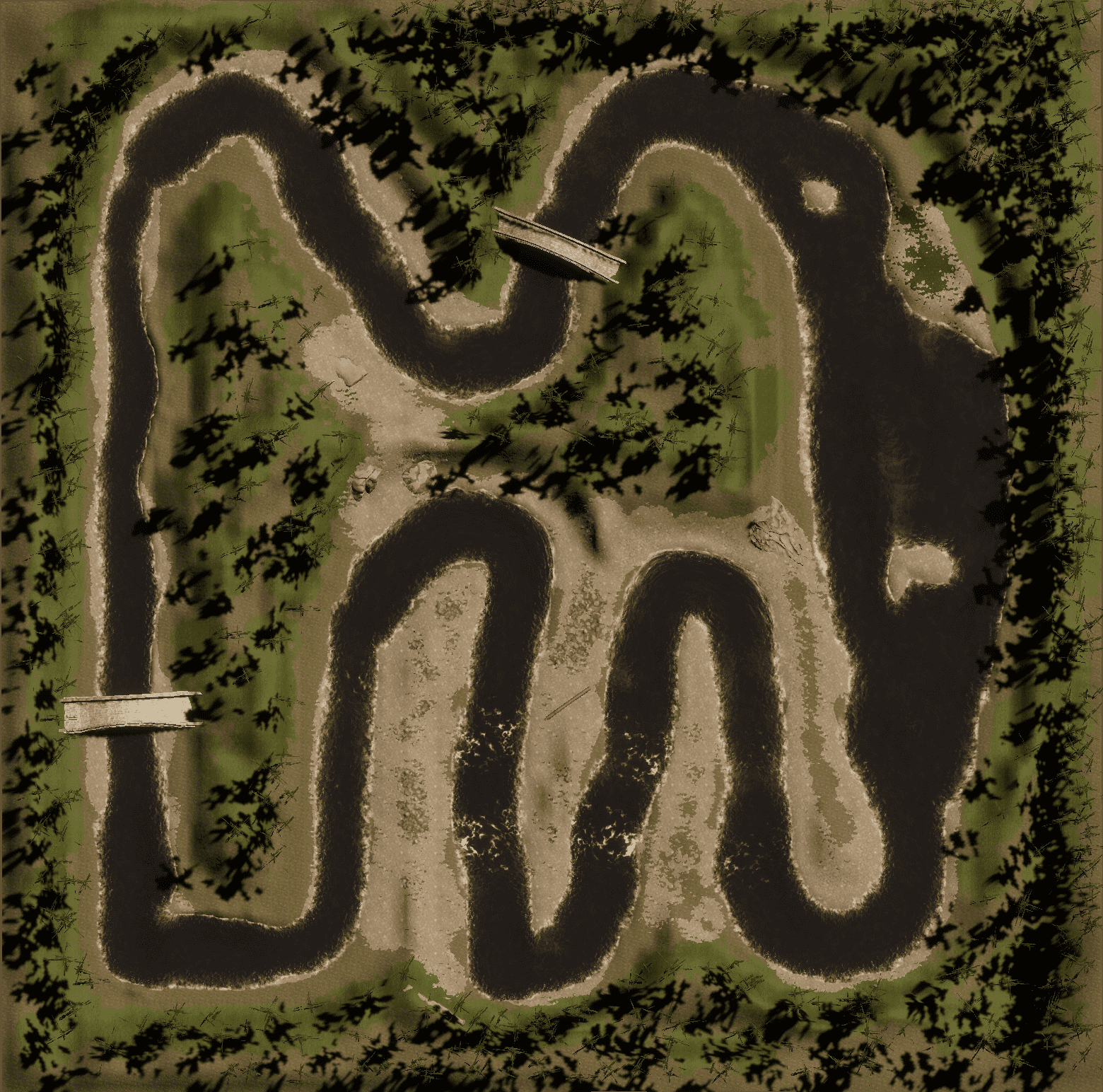}
        \caption{Hard}
    \end{subfigure}
    \caption{Orthographic view of Safe Riverine Environments of three difficulty levels}
    \label{fig:rse-bev}
\end{figure}

\revise{To facilitate the DRL training, we enhance the visual representation by reducing the dimension of Variational AutoEncoder (VAE)'s embedding, \cite{kingma2013auto}, and augmenting RGB images with semantic water masks. This encoding strategy is validated to enhance robustness in river reconstruction and provide a more informative yet compact representation of the state.}
Based on this, we benchmark several state-of-the-art Safe RL algorithms in SRE with the enhanced visual encoding method, and make in-depth comparison and analysis about the performance-safety trade-off and failure cases among these algorithms. Specifically, the performance and safety measures are compared in training phase in the \textit{Medium} environment, then in testing phase in all three environments. Orthographically viewed images of three environments in SRE are shown in Figure \ref{fig:rse-bev}.  

In summary, the main contributions of this paper are:
\begin{itemize}
    \item The presentation of Unity-based photo-realistic Safe Riverine Environment that features multiple difficulty levels and fine-grained safety metrics feedback, for training performant and safe autonomous agents in river following task.
    \item The validation of the superiority of the added semantic water mask for image encoding, as well as the choice of encoding \revise{dimension,} for more informative and compact state representation in vision-driven control.
    \item The benchmark of several state-of-the-art Safe RL algorithms in SRE for the river following task of UAV in terms of performance and safety measures during both training and testing phases, as well as detailed comparative analysis of strengths and weaknesses of these algorithms.   
\end{itemize}

 


\section{Methodology}\label{methodology}
This section gives a brief introduction to the Constrained Markov Decision Process (CMDP) for Safe RL problem in Section \ref{sec:preliminary}, and the general methods to solve the constrained optimization problem of CMDP in Section \ref{sec:algo}. 

\subsection{Preliminary}\label{sec:preliminary}
Safe RL is formalized as finding the (sub-)optimal policy in terms of reward gain while satisfying safety constraints in a CMDP, \cite{altman2021constrained}. CMDP is a tuple $(S, A, R, C, P, \mu)$, where $S$ is the set of states, $A$ is the set of actions, $R: S \times A \rightarrow \mathbb{R}$ is the reward function, $C: S \times A \rightarrow \mathbb{R}$ is the cost function, $P: S \times A \times S \rightarrow [0, 1]$ is the state transition probability function, and $\mu: S \rightarrow [0, 1]$ is the initial state distribution. Compared to Markov Decision Process (MDP) for regular RL, CMDP incorporates cost function $C$,
\revise{which serves as the penalty for undesired states and (or) actions.} 

Safe RL aims to find a policy $\pi : S \rightarrow A$ that maximizes a performance measure $J^R(\pi) \in \mathbb{R}$, while constraining a safety measure $J^C(\pi) \in \mathbb{R} \leq d$, where the policy $\pi \in \Pi$ is a mapping from states to probability distribution over actions, performance measure $J^R(\pi)$ is usually selected to be expected discounted accumulated reward over infinite horizon: $J^R(\pi) \doteq E_{\tau \sim \pi} [\sum_{t=0}^{\infty} \gamma^t R(s_t, a_t)]$, cost measure $J^C(\pi)$ is defined similarly: $J^C(\pi) \doteq E_{\tau \sim \pi} [\sum_{t=0}^{\infty} \gamma^t C(s_t, a_t)]$, and $d \in \mathbb{R}_{\geq 0}$ is a cost budget hyperparameter. Here $\tau: (s_0, a_0, s_1, a_1, ...)$ denotes the trajectory dependent on the policy $\pi$, and $\gamma$ is the discount factor. 
Thus \revise{within a feasible policy space $\Pi_{C}$, the optimal Safe RL policy} in a CMDP is    

\begin{equation}\label{eqn:cmdp}
    \pi^{\ast} = \arg \max _{\pi \in \Pi_{C}} J^R(\pi)
\end{equation}

To estimate the performance of a policy, state value function $V^{R}_{\pi} (s) \doteq E_{\tau \sim \pi} [\sum_{t=0}^{\infty} \gamma^t R(s_t, a_t) \mid s_0 = s]$ and state action value function $Q^{R}_{\pi} (s, a) \doteq E_{\tau \sim \pi} [\sum_{t=0}^{\infty} \gamma^t R(s_t, a_t) \mid s_0 = s, a_0 = a]$ are defined. $A^R_{\pi} (s, a) = Q^{R}_{\pi} (s, a) - V^{R}_{\pi} (s)$ is the reward advantage function. The value and advantage functions for \revise{$m$} costs ($V^{C_i}_{\pi} (s)$, $Q^{C_i}_{\pi} (s, a)$, $A^{C_i}_{\pi} (s, a)\revise{, \forall i \in \{1, ..., m\}}$) are defined similarly.


\subsection{Benchmarked Algorithms}\label{sec:algo}
In this section, we briefly highlight the core contributions of several state-of-the-art on-policy and off-policy model-free, and model-based Safe RL algorithms.
\revise{On-policy algorithms update policy only using experience collected by the most recent version of the policy, whereas off-policy algorithms update policy using data collected at any point during training.}
Eight model-free algorithms are selected out of all ten benchmarked algorithms due to the dynamics-free and vision-driven nature of the SRE. 
\revise{Model-free algorithms forego the potential gains of sample efficiency and online planning from using an environment model that predicts state transitions, rewards and costs.}
In order to test the potential performance improvement by learning vision-dynamics model, \revise{\cite{grigorescu2021lvd, ginerica2021vision}}, we incorporate two model-based Safe RL algorithms in our benchmark. All benchmarked algorithms, and their domains and methods, are listed in Table \ref{tab:algo}.

\subsubsection{Lagrangian Method}
converts a constraint optimization problem to an unconstrained optimization problem using adaptive penalty coefficients to enforce constraints. Given the optimization objective of CMDP in Equation \ref{eqn:cmdp}, the unconstrained problem can be formed as a Lagrangian dual problem (RCPO algorithm in \cite{tessler2018reward}) 

\begin{equation}\label{eqn:lagrangian}
    \underset{\lambda \geq 0}{\min} \hspace{1pt} \underset{\boldsymbol{\theta}}{\max} \hspace{1pt} L(\lambda, \boldsymbol{\theta}) = 
    \underset{\lambda \geq 0}{\min} \hspace{1pt} \underset{\boldsymbol{\theta}}{\max} \hspace{1pt} [J^R (\pi) - \lambda (J^C (\pi) - d)]
\end{equation}

where $L$ is the Lagrangian, $\lambda$ is the Lagrange multiplier and $\boldsymbol{\theta}$ is the parameter vector of the parameterized policy $\pi(\boldsymbol{\theta})$. 
As $\lambda$ increases, the solution to the above equation converges to that of Equation \ref{eqn:cmdp}. 
The effectiveness of the primal-dual methods is justified in \cite{paternain2022safe}, where zero duality gap is guaranteed under certain assumptions.
Both on-policy (PPO, \cite{schulman2017proximal}) and off-policy (DDPG, \cite{lillicrap2015continuous}; TD3, \cite{fujimoto2018addressing}; SAC, \cite{haarnoja2018soft}) RL algorithms can be integrated with the Lagrangian method to optimize the surrogate objective that considers both reward and costs. 
Due to the genericness of this Lagrangian primal-dual optimization method, the algorithms with suffix "Lag" refer to the original RL algorithms combined with Lagrangian method (Table \ref{tab:algo}).

\begin{table}[h]
    \centering
    \caption{Domains and methods of benchmarked Safe RL algorithms}
    \begin{tabular}{c|c|c}
        \textbf{Domains} & \textbf{Methods} & \textbf{Algorithms} \\
        \hline
        \multirow{4}{*}{On-Policy} & \multirow{1}{*}{Primal-Dual} & PPOLag \\ 
        \cline{2-3}
        & \multirow{1}{*}{Convex Optimization} & FOCOPS \\
        \cline{2-3}
        & \multirow{1}{*}{Penalty Function} & P3O \\
        \cline{2-3}
        & \multirow{1}{*}{Primal} & OnCRPO \\
        \hline
        \multirow{3}{*}{Off-Policy} & \multirow{3}{*}{Primal-Dual} & DDPGLag \\
        \cline{3-3}
        & & TD3Lag \\
        \cline{3-3}
        & & SACLag \\
        \hline
        \multirow{2}{*}{Model-based} & \multirow{2}{*}{Online Plan} & SafeLOOP \\ 
        \cline{3-3}
        & & CCEPETS \\
        \cline{2-3}
        \hline
    \end{tabular}
    \label{tab:algo}
\end{table}

\subsubsection{Constrained Policy Optimization Methods}
provide an approach for policy search in continuous CMDP that guarantees near-constraint satisfaction at each iteration, and uses trust region, \cite{schulman2015trust}, to construct surrogate functions that approximate the objectives and constraints. For parameterized stationary policies, a constrained policy optimization method tries to solve the problem

\begin{equation}\label{eqn:cpo}
\begin{split}
    \pi_{k+1} &= \arg \max_{\pi \in \Pi_{\boldsymbol{\theta}}} \underset{\substack{s \sim \d_{\pi_k} \\ a \sim \pi}}{E} [A^R_{\pi_k} (s, a)] \\
    s.t. \hspace{5pt} J^{C_i}_{\pi_k} &\leq d_i - \frac{1}{1 - \gamma} \underset{\substack{s \sim \d_{\pi_k} \\ a \sim \pi}}{E} [A^{C_i}_{\pi_k} (s, a)] \hspace{5pt} \forall i \\ 
    & \Bar{D}_{KL} (\pi || \pi_k) \leq \delta
\end{split}
\end{equation}

where $\Bar{D}_{KL} (\pi || \pi_k) = E_{s \sim \pi_k} [\Bar{D}_{KL} (\pi || \pi_k) [s]]$, $D_{KL} (\pi || \pi_k)$ is the Kullback–Leibler divergence (Equation \ref{eqn:kld}) between target policy $\pi$ and current policy $\pi_k$, and $\delta$ is the target KL divergence. 
The first constraint in Equation \ref{eqn:cpo} is derived from the objective difference between two policies (\cite{kakade2002approximately}): $J(\pi') - J(\pi) = \frac{1}{1 - \gamma} E_{s \sim \d_{\pi'}, a \sim \pi'} [A_{\pi} (s, a)]$, with the assumption of the second constraint holds so that policies within the trust region have nearly the same state distribution: $d_{\pi'} \approx d_{\pi}$. 
In this domain, First Order Constrained Optimization in Policy Space (FOCOPS, \cite{zhang2020first}) algorithm solves the above optimization objective using first-order approximation to find the optimal update policy \revise{in the non-parameterized policy space}, which is then projected to the parameterized policy space. 

\subsubsection{Model-based Online Planning Methods} employ the learning of dynamics models for better sample efficiency and faster convergence rate, compared to model-free methods, \cite{janner2019trust}. Online planning optimizes model predictive policies via the optimization of sampled trajectories and trade-off between approximated model and value functions \revise{(SafeLOOP, \cite{sikchi2022learning})}, or through cross entropy method to update policy distribution by sorting the elite policies according to their performance and feasibility in a population-based gradient-free manner \revise{(CCEPETS, \cite{wen2018constrained})}. 

Except for the above mentioned methods, several Safe RL algorithms using penalty function or primal approach are also benchmarked. Constraint-Rectiﬁed Policy Optimization (CRPO, \cite{xu2021crpo}) algorithm uses primal approach to update the policy alternatingly between objective improvement and constraint satisfaction. Penalized Proximal Policy Optimization (P3O, \cite{zhang2022penalized}) derives an equivalent unconstrained optimization problem by employing exact penalty functions instead of inclusion of dual variables, and optimizes a clipped surrogate objective, instead of a trust region objective. OnCRPO in Table \ref{tab:algo} refers to the on-policy version of CRPO.

\section{Experiments}
The components of Safe Riverine Environment as a CMDP are discussed in Section \ref{sec:sre}. The experiments of state (visual encoding) dimension and image channel dimension are described in Section \ref{sec:vae}. Benchmark experiments of Safe RL algorithms are detailed in Section \ref{sec:algo_spec}.  

\subsection{Safe Riverine Environment}\label{sec:sre}

\revise{Built on top of RSE, SRE is a Unity-based RL training environment by ml-agents toolkit, \cite{juliani2018unity}.
The UAV agent in both environments is abstracted as a camera facing $20$ degrees down.}
As a CMDP that can be abstracted by a tuple $(S, A, R, C,$
$ P, \mu)$, Section \ref{sec:preliminary}, SRE is detailed with each component of this tuple. The state space $S$ is the set of image encoding vectors by \revise{VAE}. We decide the length of the vector in accordance with the reconstruction loss of VAE networks with various latent space \revise{dimensions}. The augmentation of image channels with an additional water mask channel is also validated in virtue of Relative Entropy (RE). 
The multi-discrete action \revise{$A$} with four branches (each branch consists of actions from $\{0, 1, 2\}$) is adopted to control vertical, yaw, longitudinal and latitudinal movements of UAV \revise{in discretized manner. 
} 
SRE is dynamics-free to reduce the training difficulty, thus state transition model $P$ is deterministic. 
For initial state distribution $\mu$, we reset the UAV agent randomly in a safe pose above the circular river on episode begin. 

\revise{The safe operation space in SRE is the bounding volume lifted by the river surface.
Reward in SRE is given if the UAV agent visits some unvisited river spline segments}, the reward value is the ratio of the number of newly visited segments over all segments number, then multiplied by $10$. 
\revise{Conversely, a positive cost will be given before environment reset, if the agent travels to any unsafe pose. A safe agent stays inside the bounding volume, has no collision with bridges, and points its view close to the tangent line of the river spline.} 
In SRE, we categorize the \revise{safety} constraints into tight and loose \revise{ones}, where the former constraint violation results in $c = 1$ and the latter $c = 0.2$. Tight constraints include \{\textit{Collision}
, \textit{OutOfVolumeHorizontally}, \textit{OutOfVolumeVertically}\}, and loose constraints contain \{\textit{YawOverDeviation}, \textit{Idle}, \textit{MaxStepReached}\}
. More details can be found in \cite{wang2024vision}.

\subsection{Image Encoding}\label{sec:vae}
The length of the state vector is determined by comparing the reconstruction loss of VAE networks with different latent dimensions. 
The input image is of size $128 \times 128 \times 3$, and the reconstruction loss is the Mean Squared Error between the input image and the reconstructed image, averaged over randomly collected $2000$ image samples. 
\revise{The} overall reconstruction loss gets smaller as the latent dimension decreases, until dimension equals $8$, which is only better than $1024$-dimensional latent space network. Since the reconstruction losses for dimensions $\{64, 32, 16\}$ have no significant difference, we choose the smallest state vector size $16$ for \revise{Safe RL training}. 


We use 4-channel image input to the VAE encoder based on the intuition that the added fourth water mask channel may server as an inductive bias \revise{for better state representation and easier RL training.
On one hand, water pixels in an image provide the most related and useful information for the river following task, compared to distracting background objects like sky and terrain.
On the other hand, semantic water mask could bridge the gap between simulated and real riverine environments, \cite{wang2023aerial}, to facilitate easier Sim2Real policy transfer.
}
This \revise{4-channel encoding input} is \revise{empirically} validated by the Relative Entropy (also named Kullback–Leibler divergence) between two encoding datasets with different input image channel sizes to know straightforwardly which one contains more information than the other. Suppose $x$ belongs to some sample space $\chi$, $P(x)$ and $Q(x)$ are two discrete probability distributions of $x$, then the RE is 

\begin{equation}\label{eqn:kld}
    RE (P || Q) = \sum_{x \in \chi} P(x) \log \frac{P(x)}{Q(x)}
\end{equation}

In our experiment, $N_{sample} = 2000$ image-mask pairs are randomly collected from the \textit{medium} SRE, then separately trained using VAE with channel sizes: \{3: RGB, 1: Mask, 4: RGB+Mask\}. Encoding vector are linearly rescaled to range (-2, 2), and histogramized to probability distribution with bin number $N_{bin} = \lceil \sqrt{N_{sample}} \rceil = 45$. RE is averaged over 16 features to form Table \ref{tab:kld}.  

\begin{table}[h]
    \centering
    \caption{Relative Entropy of VAE-encoded 16-dimensional state vector of images with different channel sizes.}
    \begin{tabular}{c|c|c|c}
        \diagbox[width=\dimexpr \textwidth/8+2\tabcolsep\relax, height=0.5cm]{P(x)}{Q(x)} & RGB & MASK & RGB+MASK \\
         \hline
        RGB & - & $\boldsymbol{0.103}$ & 0.037 \\
        \hline
        MASK & 0.075 & - & 0.091 \\
        \hline
        RGB+MASK & $\boldsymbol{0.075}$ & $\boldsymbol{0.187}$ & - \\
    \end{tabular}
    \label{tab:kld}
\end{table}

\begin{figure}[h]
    \centering
    \textbf{a} \includegraphics[valign=m,height=0.8cm,width=0.42\textwidth]{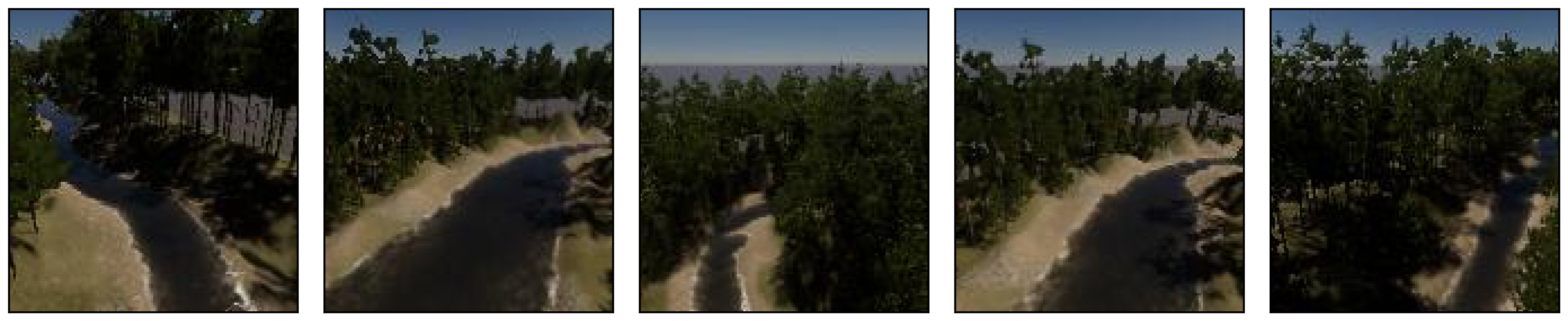} \\
    \textbf{b} \includegraphics[valign=m,height=0.8cm,width=0.42\textwidth]{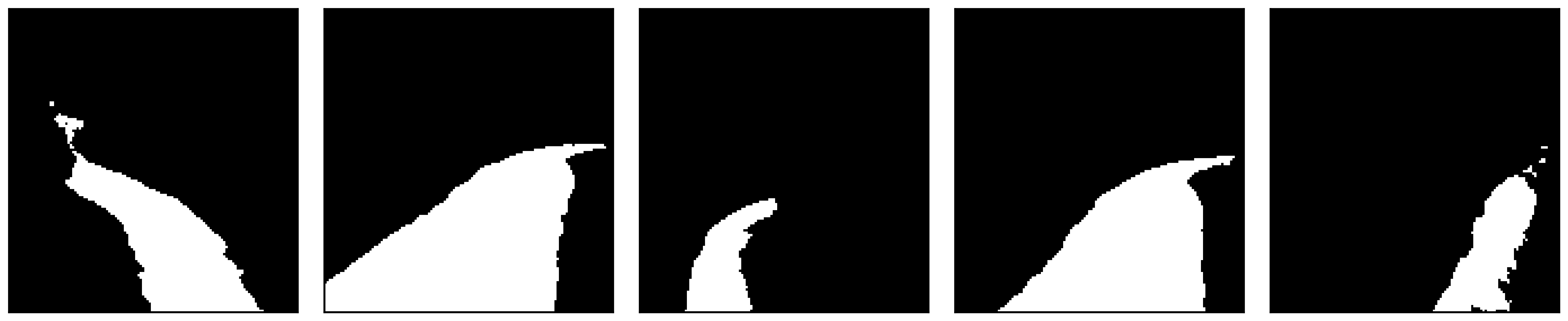} \\
    \textbf{c} \includegraphics[valign=m,height=0.8cm,width=0.42\textwidth]{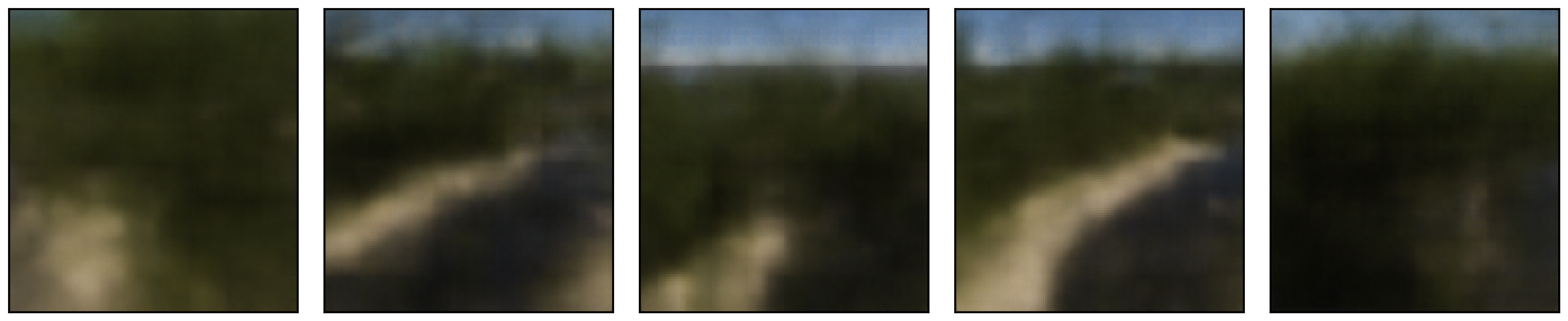} \\
    \textbf{d} \includegraphics[valign=m,height=0.8cm,width=0.42\textwidth]{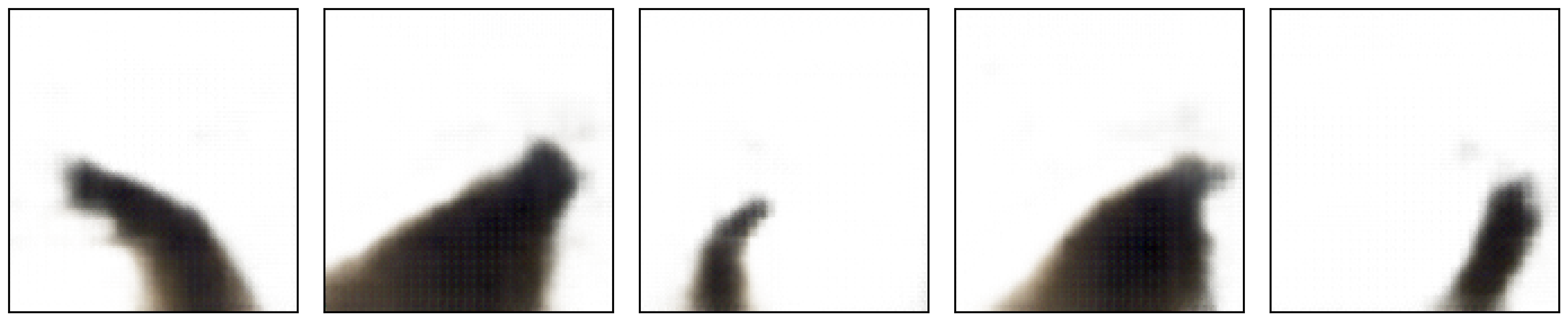} \\
    \caption{Comparison of reconstructed images of VAE trained with RGB and RGB+Mask inputs. Row \textbf{a}: source RGB images. Row \textbf{b}: source masks. Row \textbf{c}: reconstructed RGB images. Row \textbf{d}: reconstructed RGB+Mask images.}
    \label{fig:src-recon-cmp}
\end{figure}

Since RE is non-negative and non-symmetric, we measure the information gain by relative number in off-diagonal cells in Table \ref{tab:kld}. RGB+Mask encoding contains more information than both RGB and Mask encoding (bold number in last row), and expectedly, RGB encoding is informatively richer than mask encoding (bold number in first row). Qualitatively, we can validate the superiority of RGB+Mask encoding than RGB encoding in terms of the reconstruction quality of river pixels, as shown in Figure \ref{fig:src-recon-cmp}. It is obvious that some reconstructed RGB images lose the river information (row \textbf{c}), whereas all reconstructed RGB+Mask images (row \textbf{d}) retain the river channel pixels, which are an important clue for UAV to follow river based on vision input.  

\subsection{Benchmarking Specifications}\label{sec:algo_spec}
All benchmarked Safe RL algorithms are adapted from OmniSafe repository, \cite{ji2023omnisafe}. 
VAE is pre-trained with $2000$ image mask pairs \revise{randomly collected in the \textit{Medium} environment} for $100$ epochs, and the encoder part \revise{of VAE} is used \revise{in} both training and testing phases \revise{for all benchmarked} Safe RL algorithms. 
\revise{Action space shaping by discretizing diagonal Gaussian continuous action to multi-discrete action is adopted, \cite{kanervisto2020action}.}
All algorithms are trained with three seeds for $200K$ steps in the \textit{Medium} environment of SRE, and tested in all three environments \revise{with the respective seed}. 
During training phase, episodic return (averaged over last $50$ episodes) and cost rate, \cite{ray2019benchmarking}, are compared. Here cost rate is the ratio of accumulated \revise{environment reset times due to constraint violations} over current training steps.
In testing phase, algorithms are evaluated in all environments and compared according to episodic return and episodic cost, both averaged over $60$ episodes. 
PPO algorithm serves as the baseline for comparison\revise{, where only reward is used during training}.


\section{Results and Analysis}
In this section, training results in training environment
and evaluation results in all environments
are presented and analysed.

\begin{figure}[h]
    \centering
    \includegraphics[width=0.49\textwidth]{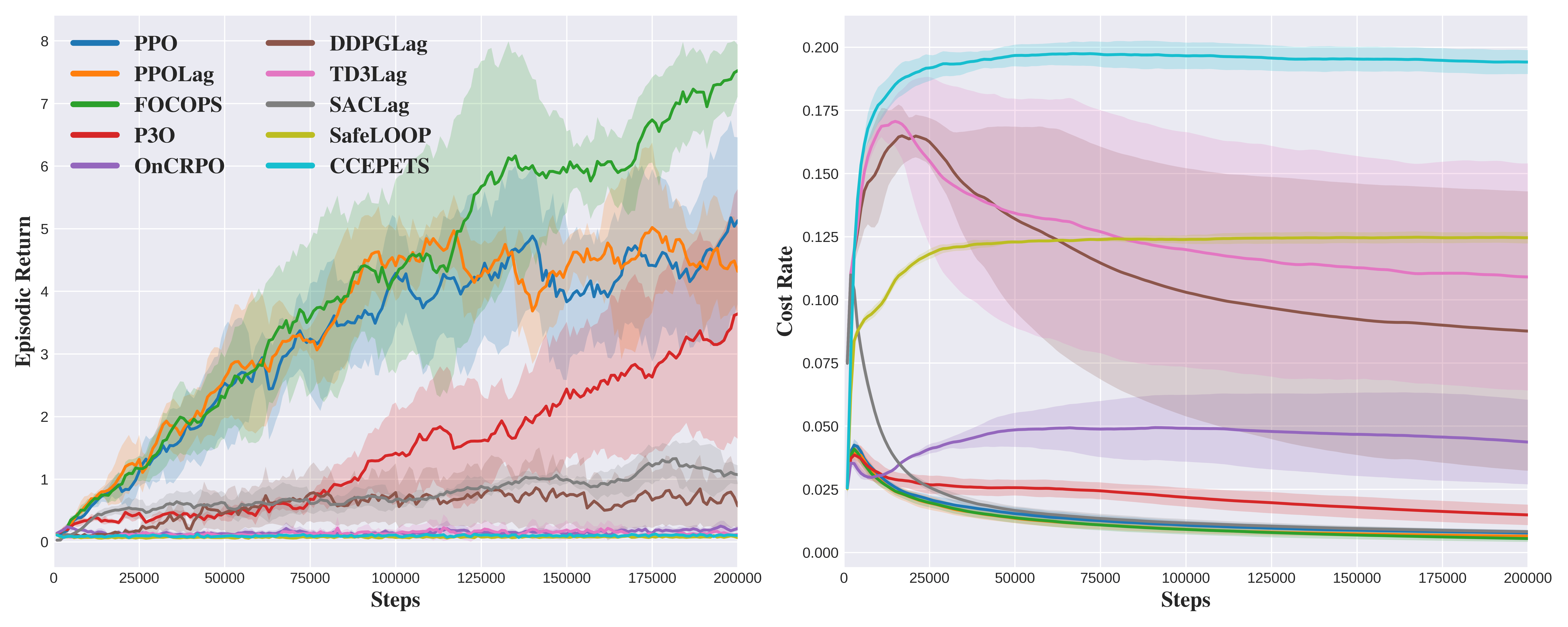}
    \caption{Episodic returns and total cost rates of Safe RL algorithms in training phase in Medium-level SRE}
    \label{fig:ret-cost}
\end{figure}

\revise{From Figure \ref{fig:ret-cost}, all off-policy algorithms (DDPGLag, TD3Lag and SACLag) and model-based algorithms 
( SafeLOOP, CCEPETS) are worse than baseline (PPO) in either episodic return or cost rate}. Specifically, model-based Safe RL algorithms have the lowest episodic return but the highest and non-decreasing cost rate, whereas off-policy algorithms have slightly better performance and overall decreasing cost rate. On the contrary, on-policy algorithms (PPOLag, FOCOPS, P3O) achieve the overall better metrics during training phase. Specifically, 
\revise{FOCOPS is the sole algorithm that exceeds the baseline in both metrics, with $1.4$ times higher episodic return, and $65\%$ cost rate of the baseline's}. 

\begin{figure}[H]
    \centering
    \includegraphics[width=0.48\textwidth]{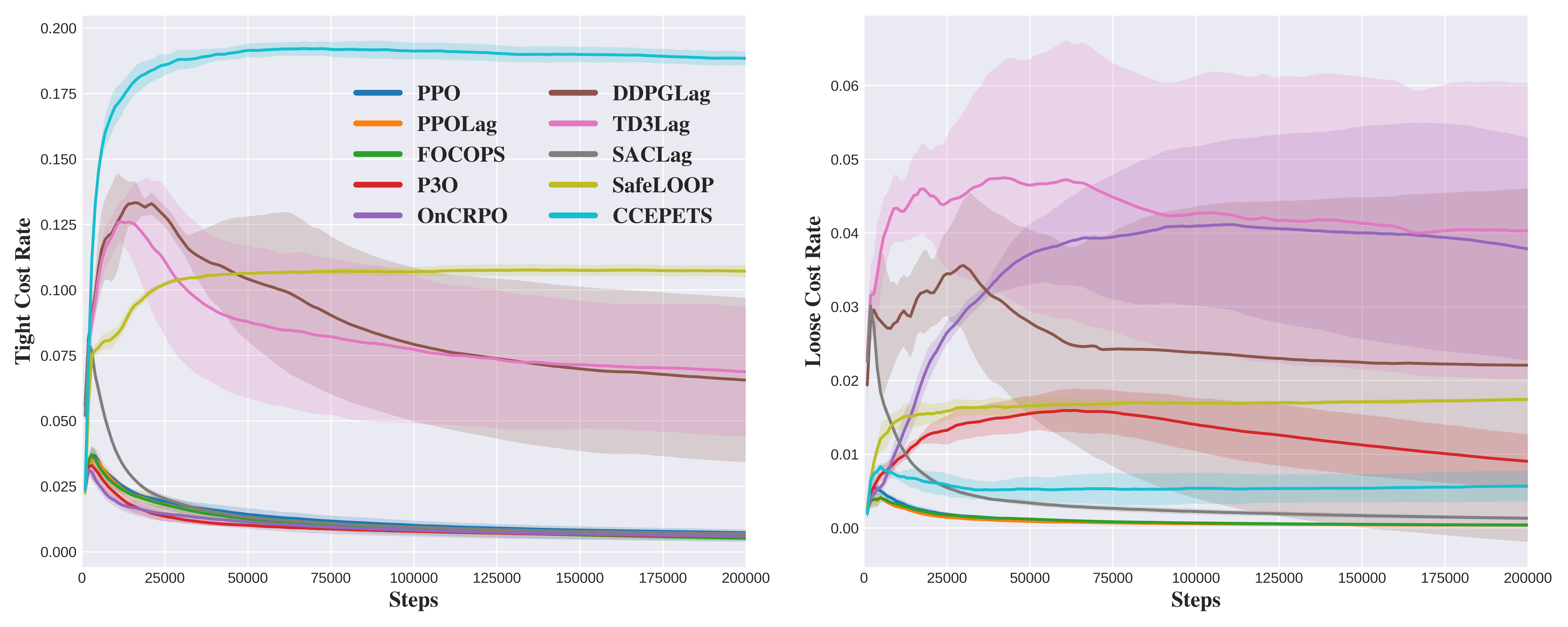}
    \caption{Tight and loose cost rates of Safe RL algorithms in training phase in Medium-level SRE}
    \label{fig:tight-loose-cr}
\end{figure}

\begin{figure*}[htb]
    \centering
    \includegraphics[height=3.3cm,width=0.98\textwidth]{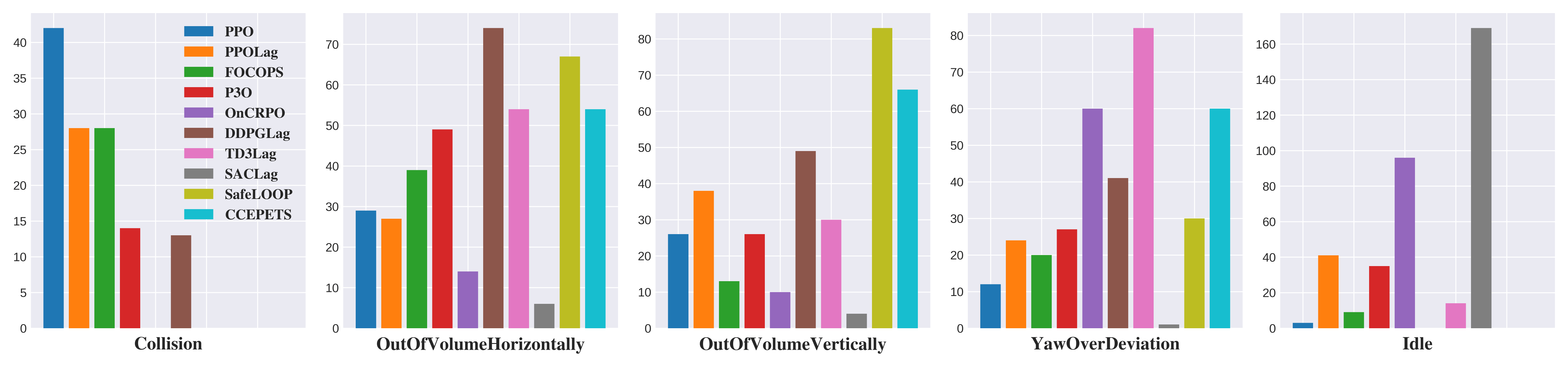}
    \caption{Aggregated failure times over $60$ evaluation episodes in three environments of $10$ benchmarked algorithms. 
    }
    \label{fig:done-reason-bar}
\end{figure*}

From Figure \ref{fig:tight-loose-cr}, tight cost rates of all algorithms have basically similar trends as the total cost rate in Figure \ref{fig:ret-cost}. 
In general, model-based algorithms have the highest violation rates of tight constraints, followed by off-policy Lagrangian algorithms, then on-policy algorithms. 
Combined with the loose cost rate figure, OnCRPO algorithm makes more soft constraints violations in terms of soft violation percentage over all violations, meaning the agent often triggers less severe constraints by making consecutive non-progressive actions in situ. The same phenomenon can also be observed in Figure \ref{fig:done-reason-bar}. 
This is inline with the feature of CRPO: immediate switches between optimizing the objective and reducing the constraints whenever they are violated. However, the exploration ability is limited in this primal-based algorithm, compared to the other Lagrangian-based ones.

\begin{table*}[htb]
\centering
\caption{Evaluation results of Safe RL algorithms (trained in \textit{Medium} SRE) tested in three levels of SRE maps in terms of episodic return (EpR) and episodic cost (EpC) averaged over 60 episodes (20 episodes per seed)}
\begin{tabular}{c|c|c|c|c|c|c}
\hline
Env & \multicolumn{2}{l|}{  \hspace{40pt}  Easy} & \multicolumn{2}{l|}{\hspace{35pt} Medium} & \multicolumn{2}{l}{\hspace{40pt} Hard} \\ \hline
\diagbox[width=\dimexpr \textwidth/8+\tabcolsep\relax, height=0.5cm]{Algo}{Metrics} & EpR & EpC & EpR & EpC & EpR & EpC \\ \hline 
PPO (baseline) & $9.19 \pm 2.63$ & $0.09 \pm 0.28$ & $5.22 \pm 3.89$ & $0.74 \pm 0.44$ & $1.34 \pm 1.34$ & $0.84 \pm 0.32$ \\ \hline \hline
PPOLag & $5.86 \pm 3.85$ & $0.32 \pm 0.38$ & $3.81 \pm 3.36$ & $0.79 \pm 0.37$ & $0.96 \pm 0.87$ & $0.65 \pm 0.40$ \\ \hline
FOCOPS & $\boldsymbol{7.83 \pm 3.74}$ & $\underline{\boldsymbol{0.15 \pm 0.32}}$ & $\boldsymbol{6.30 \pm 4.41}$ & $0.49 \pm 0.49$ & $\boldsymbol{2.05 \pm 1.96}$ & $0.79 \pm 0.35$ \\ \hline
P3O & $6.23 \pm 3.93$ & $0.37 \pm 0.44$ & $3.14 \pm 3.43$ & $0.67 \pm 0.41$ & $0.85 \pm 0.82$ & $0.65 \pm 0.40$ \\ \hline
OnCRPO & $0.27 \pm 0.50$ & $0.27 \pm 0.22$ & $0.10 \pm 0.14$ & $0.29 \pm 0.26$ & $0.15 \pm 0.23$ & $0.36 \pm 0.32$ \\ \hline
DDPGLag & $1.75 \pm 2.55$ & $0.78 \pm 0.37$ & $0.73 \pm 0.91$ & $0.83 \pm 0.33$ & $0.44 \pm 0.64$ & $0.80 \pm 0.35$ \\ \hline
TD3Lag & $0.14 \pm 0.12$ & $0.55 \pm 0.40$ & $0.13 \pm 0.10$ & $0.69 \pm 0.39$ & $0.10 \pm 0.10$ & $0.48 \pm 0.38$ \\ \hline
SACLag & $0.06 \pm 0.18$ & $0.20 \pm 0.00$ & $0.02 \pm 0.03$ & $\underline{\boldsymbol{0.27 \pm 0.22}}$ & $0.04 \pm 0.11$ & $\underline{\boldsymbol{0.27 \pm 0.22}}$ \\ \hline
SafeLOOP & $0.12 \pm 0.12$ & $0.87 \pm 0.30$ & $0.12 \pm 0.11$ & $0.92 \pm 0.24$ & $0.15 \pm 0.24$ & $0.84 \pm 0.32$ \\ \hline
CCEPETS & $0.08 \pm 0.10$ & $0.79 \pm 0.35$ & $0.05 \pm 0.04$ & $0.79 \pm 0.35$ & $0.08 \pm 0.23$ & $0.69 \pm 0.39$ \\ \hline
\end{tabular}%
\label{tab:test}
\end{table*}

Table \ref{tab:test} lists the evaluation results of agents trained in \textit{Medium} SRE and evaluated in all three environments in terms of episodic return and episodic cost. In SRE, the maximum attainable episodic return is $10$, and the maximum episodic cost is $1$ 
. 
It is notable that FOCOPS achieves the highest episodic return (bold) in all environments, and the lowest episodic cost (bold with underline) in \textit{Easy} environment. 
FOCOPS achieves better results than baseline in both \textit{Medium} and \textit{Hard} environments, only worse than baseline in \textit{Easy} environment. 

Detailed failure case visualization is presented in Figure \ref{fig:done-reason-bar}, where the total number of failures for each failure reason give intuitive impression of how algorithms behave in terms of safety. 
\revise{No algorithm triggers $\textit{MaxStepReached} = 500$ failure since each river needs less than $500$ perfect steps to traverse, thus \textit{Success} or \textit{Idle} failure will be triggered beforehand.}
First of all, baseline PPO has almost linearly decreasing failure numbers as the failure severity decreases (from left to right subplot in Figure \ref{fig:done-reason-bar}), making it a decent baseline algorithm which focuses only on task objective without safety consideration. The only algorithm resembling this trend is FOCOPS, with overall lower failure cases. The only algorithm that reverses this trend is OnCRPO, which has lower failure times than baseline in tight constraints, but higher violations of loose constraints. 
Secondly, the model-based algorithms have trouble in making continuous valid safe actions so that the collision with bridges or being idle is not triggered at all. 
Thirdly, on-policy algorithms have overall lower number of failures than off-policy and model-based algorithms, which is also validated in Figure \ref{fig:ret-cost} and Table \ref{tab:test}. In this regard, we speculate that the visual dynamics and characteristics in SRE affect the effectiveness of off-policy correction methods like Importance Sampling (IS), thus cause data distribution mismatch, breaking the assumption of off-policy algorithms. 

In summary, SRE is shown to be a difficult Safe RL environment for agent to operate in nicely and safety. Several findings can be highlighted based on the benchmarking results to inspire the future work. 
First, the scale and value of the cost as a penalty term in Lagrangian methods need to be investigated since the unclipped increasing Lagrange multiplier may limit the exploration ability of the policy. 
Second, immediate response to constraints violation (CRPO) provides good safety compliance, whereas the balance of it with reward maximization needs more attention. 
Third, the discretization of action space may cause strongly IS-dependent off-policy algorithms to struggle, thus adaptation of SRE to continuous action space needs to be experimented with.
Fourth, the effectiveness of learned visual dynamics in model-based algorithms needs to be researched to make sure the dynamics model is actually helping safe policy learning instead of harming it. 
Fifth, FOCOPS shows good balance between exploration and safety, proving the benefits of optimization in unparameterized policy space then projection into the parameterized one, as well as the simple first order approximation. 
However, based on its safety violation trend in Figure \ref{fig:done-reason-bar}, how to better discriminate between tight and loose constraints to train safer policy remains a problem for this algorithm.  

\section{Conclusion}
In conclusion, this paper makes several contributions to the field of safe vision-driven \revise{autonomous navigation in confined waters}.
The development of a Unity-based photo-realistic Safe Riverine Environment (SRE) introduces multiple difficulty levels and provides fine-grained safety metrics feedback, enabling the \revise{RL} training of autonomous agents for both performance and safety in river following tasks. 
Additionally, the study validates the effectiveness of incorporating a semantic water mask for image encoding and explores the optimal encoding \revise{dimension}, enhancing the informativeness and compactness of state representation in vision-driven control.
Furthermore, the paper conducts a comprehensive benchmark of several state-of-the-art Safe RL algorithms within the SRE, evaluating \revise{and analysing} their performance and safety measures, \revise{in terms of both tight and loose constraint violations}. 
\revise{Notably, the proposed vision encoding method is also applicable to vision-driven navigation of ASVs in rivers, and the benchmarking results of Safe RL algorithms offer valuable insights for safe autonomy in maritime domain.}


\bibliography{ifacconf}

\end{document}